\documentclass[12pt]{iopart}
\usepackage[export]{adjustbox}
\usepackage{multirow}%
\usepackage{amsmath,amssymb,amsfonts}%
\usepackage{amsthm}%
\usepackage{mathrsfs}%
\usepackage[title]{appendix}%
\usepackage{textcomp}%
\usepackage{manyfoot}%
\usepackage{booktabs}%
\usepackage{algorithm}%
\usepackage{algorithmicx}%
\usepackage{algpseudocode}%
\usepackage{listings}%
\usepackage{subcaption}
\usepackage{hyperref} 

\begin{document}

\title[Real-Time Sequential Facial Expression Analysis Using Geometric Features]{Deep Learning-Based Real-Time Sequential Facial Expression Analysis Using Geometric Features}

\author{Talha Enes Köksal}
\ead{talhaeneskoksal@gmail.com}
\address{Electrical and Electronics Engineering, Izmir Institute of Technology, Türkiye}
\author{Abdurrahman Gumus}
\ead{abdurrahmangumus@isparta.edu.tr}
\address{Computer Engineering, Isparta University of Applied Sciences, Türkiye}


\begin{abstract}
Facial expression recognition is a crucial component in enhancing human-computer interaction and developing emotion-aware systems. Real-time detection and interpretation of facial expressions have become increasingly important for various applications, from user experience personalization to intelligent surveillance systems. This study presents a novel approach to real-time sequential facial expression recognition using deep learning and geometric features. The proposed method utilizes MediaPipe FaceMesh for rapid and accurate facial landmark detection. Geometric features, including Euclidean distances and angles, are extracted from these landmarks. Temporal dynamics are incorporated by analyzing feature differences between consecutive frames, enabling the detection of onset, apex, and offset phases of expressions. For classification, a ConvLSTM1D network followed by multilayer perceptron blocks is employed. The method's performance was evaluated on multiple publicly available datasets, including CK+, Oulu-CASIA (VIS and NIR), and MMI. Accuracies of 93\%, 79\%, 77\%, and 68\% were achieved respectively. Experiments with composite datasets were also conducted to assess the model's generalization capabilities. The approach demonstrated real-time applicability, processing approximately 165 frames per second on consumer-grade hardware. This research contributes to the field of facial expression analysis by providing a fast, accurate, and adaptable solution. The findings highlight the potential for further advancements in emotion-aware technologies and personalized user experiences, paving the way for more sophisticated human-computer interaction systems. To facilitate further research in this field, the complete source code for this study has been made publicly available on GitHub: \url{https://github.com/miralab-ai/facial-expression-analysis}.
\end{abstract}


\vspace{2pc}

\maketitle

\section{Introduction}\label{sec1}

Humans possess the unique ability to communicate emotions through their facial expressions, which are considered one of the most powerful, natural, and universal forms of expression \cite{li2020deep}. Connection between emotions and facial expression was found in a collaborative study with Ekman, Levenson and Friesen. It is discovered that performing certain facial muscular actions generates emotion physiology \cite{ekman1992facial}. In 1972, Ekman et al. conducted a review of prior research on the interpretation of facial expressions in western cultures and discovered that all studies found evidence of six basic emotions: happiness, surprise, fear, sadness, anger, and disgust with a hint of contempt. They observed that in some cultures, fear and surprise can be identical and hard to classify \cite{ekman1992there}.
\par
The analysis of facial expressions relies on the extraction of specific features, typically categorized into two types according to their feature representations: spatial and spatio-temporal \cite{li2020deep}. Spatial features represent information derived from static images, like a single photo capturing a distinct facial expression that reveals a specific emotion such as joy, anger, or surprise. On the other hand, spatio-temporal features encapsulate data extracted from a series of images, akin to a video that sequences a person’s emotional display over time \cite{Chen2023, CHEN2024123635}. This method captures the dynamic progression of facial movements, offering a comprehensive and nuanced insight into how a person’s emotional state might evolve. For instance, it could trace the transition from surprise to delight, or from calm to fury, which helps to track the temporal development and complexity of emotions \cite{pantic2006dynamics}.
\par
The process of facial expression recognition task includes four main steps \cite{sharma2019automatic}. The first of these is pre-processing, which includes detecting the face within an image or a series of images. Once the face is detected, the process advances to the second step: generating additional facial content, such as identifying and mapping the facial landmarks. These landmarks, which include key features like the eyes, nose, mouth, and contour of the face, provide a detailed facial structure that becomes instrumental in the subsequent stage of feature extraction. The third step is feature extraction, where the system isolates important attributes from the face using the generated landmarks. This captures the unique aspects of each facial expression and prepares the data for the final stage. In the concluding step, emotion classification, the system interprets the extracted facial features, assigning them to specific emotional states, which could range from basic emotions like happiness, sadness, or anger, to more complex emotional nuances \cite{xie2022overview}.
\par
There are three commonly used techniques in facial feature extraction that are prevalent in the literature: geometric, appearance-based, and motion-based methods \cite{kumari2015facial, mollahosseini2016going}. Appearance- based methods, one of the foremost approaches, utilizes a pixel-based approach to extract facial features. State-of-the-art techniques often incorporate attributes such as pixel intensities, Gabor filters, Local Binary Patterns (LBP) \cite{Saleem_2021}, Local Phase Quantization (LPQ) \cite{Karanwal_2022}, and Histogram of Oriented Gradients (HoG) \cite{Chang2019} to obtain information about the face \cite{mollahosseini2016going}. Meanwhile, motion-based methods focus on characteristics related to movement, such as shifts in position and shape. These alterations are predominantly driven by the contractions and relaxations of facial muscles during emotional expressions \cite{zhang2011facial}. Techniques in this domain might involve optical flow, Motion History Images (MHI), and volume LBP to capture these dynamic changes \cite{mollahosseini2016going}.
\par
In feature extraction perspective, Convolutional Neural Network (CNN) which is widely used by deep learning frameworks also utilized as the most common methodology to extract features for pixel centric approaches \cite{aloysius2017review, Elsheikh2025, MANAVAND2025121658, SHAHID2023110451}. Geometric features that are extracted with the help of landmarks are often mathematical attributes like Euclidean distance, slope, angle and coordinates of landmarks, landmark trajectories. \cite{alvarez2018facial, buhari2020facs, qiu2019facial, rohith2020facial, sharma2019automatic, Kumar2025}.
\par
One of the key advantages of geometric features is their ease of computation, as they require relatively less processing power compared to more complex feature extraction methodologies such as CNNs\cite{Wang2023}. This results in faster processing of frames in a sequence, making geometric features an attractive choice for real-time applications. Furthermore, geometric features are highly robust to unwanted disruptions in the facial image, such as variations in illumination, rotation, and misalignment. These disruptions can often be found conventional pixel-centric and motion-based features, leading to reduced accuracy and reliability in facial expression recognition tasks. By contrast, geometric features are able to circumvent such disruptions by focusing on the underlying structure of the face, resulting in more accurate and reliable recognition of facial expressions.
\par
Given the critical role of facial expressions in understanding emotions \cite{ekman1992facial}, the study of emotion is highly dependent on the measurement of facial expressions, leading to the development of several observer-based systems. Among these systems, the Facial Action Coding System (FACS) stands out as the most extensively used and recognized for its comprehensive methodology, psychometric rigor, and broad applicability across diverse scenarios \cite{cohn2007observer}. FACS is the most commonly utilized scheme for breaking down facial expressions into their individual muscle movements, referred to as Action Units (AUs). FACS enables the description of any facial expression as a combination of specific Action Units (AUs) providing a systematic approach for analyzing and understanding the complexities of facial expressions. Ekman and Friesen initially proposed the FACS and later updated it in 2002 to account for micro-expressions \cite{xie2022overview}.
\par
Macro-expressions are commonly observed during daily interactions. Typically, lasting between 0.5 to 4 seconds, they manifest with noticeable visibility and intensity. In contrast, micro-expressions are fleeting, existing for no longer than half a second and can easily be overlooked without focused attention. This duration and intensity differentiate these two types of expressions. Generally, macro-expressions present themselves with higher visibility and intensity than micro-expressions, making them easier to recognize \cite{xie2022overview}. 
\par
The process of emotional expression in the face can be categorized into three consecutive temporal phases: onset, apex and offset. Onset is the starting phase of emotional expression, where the first hints of an emotion begin to surface. Next, the emotional expression escalates to the apex phase. Here, the emotion is fully visible, reaching its peak intensity. This is the stage where the emotion is most pronounced and easily identifiable. The final phase is offset, characterized by a gradual relaxation of the facial muscles post-apex. In this stage, the intensity of the emotional expression slowly diminishes, signaling the end of the emotional display \cite{wu2014survey}.
\par
In this paper, we present real-time sequential macro expression recognition method using geometric features extracted from facial landmarks. We begin with a literature review of existing methods for macro facial expression recognition, with a particular focus on geometric features. We then describe our methodology for real-time sequential macro expression recognition, which includes facial landmark detection, feature extraction, and classification using machine learning techniques. We present experimental results and provide a comprehensive discussion regarding the performance of our approach in terms of recognition accuracy, processing speed, and robustness in handling variations in facial expression intensity. Finally, we conclude with a summary of our contributions and future directions for research in this area.

\section{Background and Literature Review}\label{sec2}

In this study, we employed a landmark-based facial feature extraction approach, deviating from the more commonly utilized appearance-based (pixel centric) methodology prevalent in the literature. Here, several prominent landmark-based studies that have contributed significantly to facial expression recognition are highlighted. Choi et al. adopted sequential approach for representing facial features, employed facial landmarks to calculate the distances between all points and deriving their differences for consecutive frames, produced construct termed as Landmark Feature Maps (LFM). These LFMs were normalized to a range 0-255, generating LFM images. The facial features for each LFM were then extracted using a VGG13-based Convolutional Neural Network (CNN). The final layer incorporated Long Short-Term Memory (LSTM) and Multilayer Perceptron (MLP) for the classification  \cite{choi2020facial}. Building on the LFM methodology, Kim et al. proposed “squeezed LFM” designed to eliminate redundant duplicate data within the LFM. They noted the inherent symmetry of LFMs about a diagonal axis, given that the distance from point x to point y mirrors that from point y to point x \cite{kim2021facial}. Apart from comparing distances between two points, alternate distance feature approaches have been proposed. For instance, Raj et al. proposed identifying a central point by calculating the mean of both axes, followed by determining the distance of all points relative to this central point \cite{rohith2020facial}. Meanwhile, Alvarez et al. fed a multilayer perceptron with two inputs: the first being the facial landmark coordinates of a person showing an emotion, and the second being the distance between this coordinates and the neutral state landmark coordinates of the same individual \cite{alvarez2018facial}. Similarly, Sharma et al. proposed raw landmark coordinates and certain Euclidian distances that are manually picked as features \cite{sharma2019automatic}. Qui and Wan proposed to create an input vector by subtracting all landmark points relative to their regional center points. They divided the face into four regions, each with its own center point \cite{qiu2019facial}. Beh et al. proposed six Euclidian distances selected manually over eyebrow, eye and mouth regions of face. The ratios of these distances to a reference distance were selected as features \cite{beh2019micro}. Khan et al. proposed all Euclidian distances of each pair of extracted landmarks and additionally Euclidian distance of all landmarks relative to average point on the face to be used as features \cite{khan2018facial}. Buhari et al. proposed both Euclidian distances and slopes of landmark pairs to be used as features. Facial regions were created based on Facial Action Coding System proposed by Paul Ekman. Features extracted from full face landmarks and landmarks belongs to created regions were experimented separately \cite{buhari2020facs}.
In the literature mostly Dlib library’s pre-trained facial landmark detector is used to extract facial landmarks \cite{alvarez2018facial, beh2019micro, buhari2020facs, rohith2020facial}. There are also several other algorithms like incremental Parallel Cascade of Linear Regression (iPar–CLR) \cite{sharma2019automatic} and landmark detector of IntraFace software package \cite{khan2018facial}. In the study by Choi et al, facial landmark detection method which is called Supervision-by-Registration (SBR) is used \cite{choi2020facial}. Also, some datasets like the Extended Cohn-Kanade Dataset (CK+) comes with ready to use landmarks along with it which is tracked by an Active Appearance Model algorithm \cite{lucey2010extended}. In preprocessing perspective, pixel centric approaches often require preprocessing before feeding image into neural network. These can be face alignment, scaling, rotation, illumination and color fixes, background and noise removal \cite{li2020deep}. Landmark based approaches do not require additional preprocessing as long as landmark detection algorithm can detect required landmarks since all preprocessing is handled by the algorithm itself.

In this study, MediaPipe Face Mesh is used to detect facial landmarks since it has an impressive performance on GPUs and can deliver 478 landmarks in total that is higher than other landmark detection methods \cite{bazarevsky2019blazeface}. To evaluate the relative performance of MediaPipe FaceMesh and the other popular package dlib, the average processing time of both algorithms on a subject is measured. The methodology involves a multi-step process. Firstly, the entire sequence of camera frames capturing facial expressions is reduced to five key frames, representing the transition from a neutral expression to the apex of the expression. These selected frames are then processed using the MediaPipe FaceMesh Solution to extract the positions of facial landmarks accurately. For each chosen pair of landmarks, both the Euclidean distance and angle features are calculated. To capture temporal dynamics, the differences between the current frame features and the corresponding features of the previous frame are computed. Finally, these two feature types are concatenated to create a comprehensive feature vector. For real-time detection, a sliding window buffer that holds five frames is used; when the buffer reaches five, the pre-trained model takes the buffer and predicts the emotion.
No preprocessing is introduced before using images from datasets, and only raw images are used. Additionally, no data augmentation is applied to increase subject count.

The rest of the paper is organized as follows: In Section 3, Methodology of the study such as computational setup, used datasets, feature creation and classification algorithms are discussed. In Section 4 Results and Discussions such as experiments with datasets individually, composite dataset experiments and real-time implementation are discussed.

\section{Methodology}

\subsection{Computational Setup}
Our experiments are conducted in the environments with following specs: Ubuntu OS, Intel i5-12600K CPU, 64 GB RAM, NVIDIA GeForce RTX 3060 12 GB GPU.
Training of the proposed framework and the preprocessing operations have functioned using the Python programming language. Model is implemented in the Keras backend of the TensorFlow 2.1 framework. The categorical cross-entropy loss function and the Adam optimizer with default settings are used during the training. The batch size and number of epochs are selected as 32 and 200. 

\subsection{Datasets}
\begin{figure}[htbp]
  \centering
  \includegraphics[width=\textwidth]{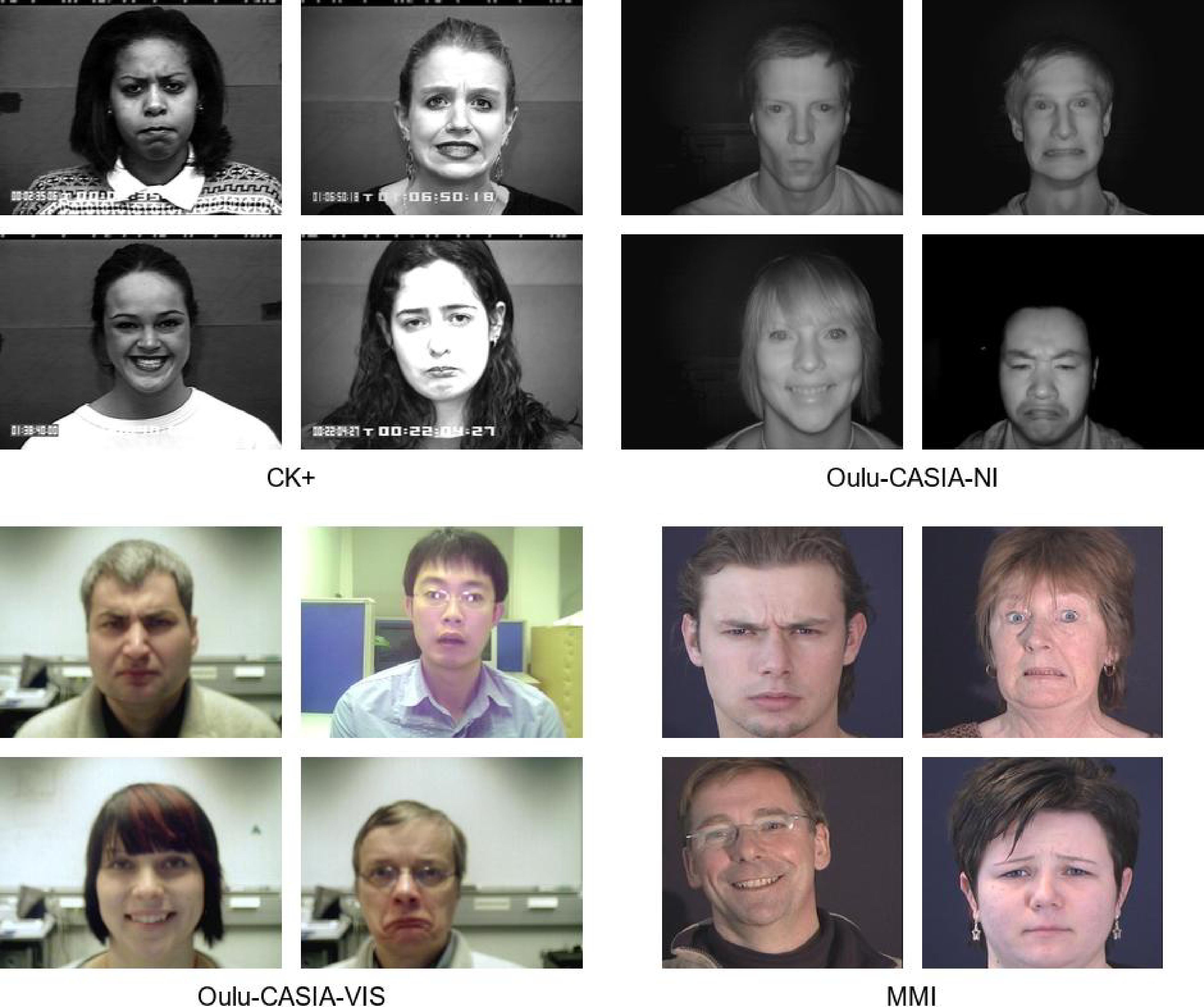}
  \captionsetup{format=hang}
  \caption{Collage of images in datasets which are used in macro-expression experiments. (a) Cohn-Kanade (CK+) dataset. (b) Oulu-CASIA dataset Near Infrared (NI) variation. (c) Oulu-CASIA dataset Visible Light (VIS) variation. (d) MMI dataset, only subjects that have sequential data is used.}
  \label{fig:collage_macro}
\end{figure}

Given the focus of this study on the spatio-temporal features, only sequential datasets are employed for the analysis. Figure \ref{fig:collage_macro} shows example images of subjects from datasets and Table \ref{tab:dataset_table} gives summary of used datasets.
\par
\begin{table}[htbp]
    \centering
    \small
    \caption{Summary of datasets used in macro expression experiments.}
    \label{tab:dataset_table}
    \setlength\tabcolsep{2.2pt}
    \begin{tabular}{lcccccc}
        \hline
        \multirow{2}{*}{\textbf{Dataset}} & \textbf{Subject} & \textbf{Sequence} & \multirow{2}{*}{\textbf{FPS}} & \multirow{2}{*}{\textbf{Resolution}} & \textbf{Emotion} & \multirow{2}{*}{\textbf{Ethnicity}} \\
        & \textbf{Count} & \textbf{Count} & & & \textbf{Count} & \\
        \hline
        \multirow{3}{*}{CK+} & \multirow{3}{*}{123} & \multirow{3}{*}{593} & \multirow{3}{*}{30} & \multirow{3}{*}{640x480} & \multirow{3}{*}{7} & Euro-American (81\%) \\
        & & & & & & Afro-American (13\%) \\
        & & & & & & Other groups (6\%) \\
        & & & & & & \\
        \multirow{2}{*}{Oulu-CASIA} & \multirow{2}{*}{80} & \multirow{2}{*}{2472} & \multirow{2}{*}{25} & \multirow{2}{*}{320x240} & \multirow{2}{*}{6} & Finnish (\textasciitilde60\%) \\
        & & & & & & Chinese (\textasciitilde40\%) \\
        & & & & & & \\
        \multirow{3}{*}{MMI} & \multirow{3}{*}{19} & \multirow{3}{*}{848} & \multirow{3}{*}{24} & \multirow{3}{*}{720x576} & \multirow{3}{*}{6} & European \\
        & & & & & & Asian\\
        & & & & & & South American\\ 
        \hline
    \end{tabular}
\end{table}

The CK+ is a fully FACS-compatible dataset with 593 sequences captured from 123 subjects at 30 frames per second (FPS) with either 640x490 or 640x480 pixels resolution. With the aid of two precisely synchronized Panasonic AG-7500 cameras, the facial expressions of 210 individuals were meticulously captured and analyzed. The participants, who ranged in age from 18 to 50, were predominantly female (69\%) and of Euro-American descent (81\%), with Afro-American (13\%) and other groups (6\%) making
up the remainder. Under the guidance of an experimenter, the subjects were directed to execute a series of 23 facial expressions, which encompassed both individual action units and various combinations thereof. Each sequence starts from a neutral state, ends at the apex phase, and captures duration; hence frame count is different for each sequence. The apex frame of each sequence is validated and labeled by emotion researchers with reference to FACS Investigators Guide. The dataset consists of seven emotions, namely anger, contempt, disgust, fear, happiness, sadness, and surprise \cite{kanade2000, lucey2010extended}. The Oulu-CASIA is a sequential dataset with two variations: one is captured with visible light conditions (VIS), and the other one is captured with near- infrared conditions (NIR). A total of 80 people between 23 and 58 years old participated, and all expressions were captured at 25 FPS with an image resolution of 320×240 pixels. The database is comprised of two distinct parts. The first segment was captured in February 2008 by the Machine Vision Group of the University of Oulu in Finland, featuring a total of 50 subjects, a majority of whom were Finnish individuals. The second portion was recorded in April 2009 in Beĳing by the National Laboratory of Pattern Recognition, Chinese Academy of Sciences, and consisted of 30 subjects who were all of Chinese descent. Participants were instructed to take a seat in front of the camera in an observation room, with a distance of approximately 60 cm between their face and the camera. They were then prompted to imitate a facial expression as demonstrated in a series of pictures. Images with three different illumination conditions: weak, normal, and dark, are present in the dataset. Each sequence starts from a neutral state and ends at the apex phase \cite{zhao2011facial}. The MMI Face Database is a complex source that contains both static and sequential images captured at frontal and profile views of faces. Every video sequence in the database was captured at a standard rate of 24 frames per second using a PAL camera. The collection comprises roughly 30 profile-view and 750 dual-view facial expression video sequences. These sequences differ in length, ranging from 40 to 520 frames, and portray one or multiple facial behavior patterns, starting with a neutral facial expression, followed by an expressive one, and ending with another neutral expression. The database features 19 distinct faces, belonging to both male and female students and research staff members, with an ethnic background of either European, Asian, or South American. The total number of female faces is 4400. The ages of the participants range from 19 to 62 years old. They were directed by a FACS coder on how to execute 79 different series of expressions and were asked to include a brief neutral state at the beginning and end of each expression. Onset apex and offset phases can be studied for this database \cite{pantic2005web, Valstar2010InducedD}.
\subsection{Feature Creation Algorithm}
Facial landmarks used in this paper provide a basis for deriving geometric features. These landmarks should be accurately positioned, and detection should be fast enough to achieve real-time performance for facial expression recognition tasks. For these reasons, MediaPipe FaceMesh solution is used as a facial landmark detector. MediaPipe FaceMesh is a facial landmark detection solution developed by Google’s MediaPipe team. It uses machine learning to identify and track 478 facial landmarks on a person’s face, including the eyes, eyebrows, nose, mouth, and jawline \cite{kartynnik2019real}. MediaPipe utilizes a lightweight and very fast, 200-1000 FPS on mobile GPUs, face detector, which is called BlazeFace \cite{bazarevsky2019blazeface}. The face landmark model is a neural network-based model that estimates 478 landmarks with 3D coordinates. It uses a single camera output frame as an input to the model. This model is lightweight and applicable for real-time tasks with 100-1000 FPS on mobile GPUs [25]. Attention mesh is an optional step that applies attention to the eye, iris, and lip regions. As a result, estimated landmarks are more accurate on these regions \cite{grishchenko2020attention}.
\par
\begin{figure}[htbp]
  \centering
  \includegraphics[width=\textwidth]{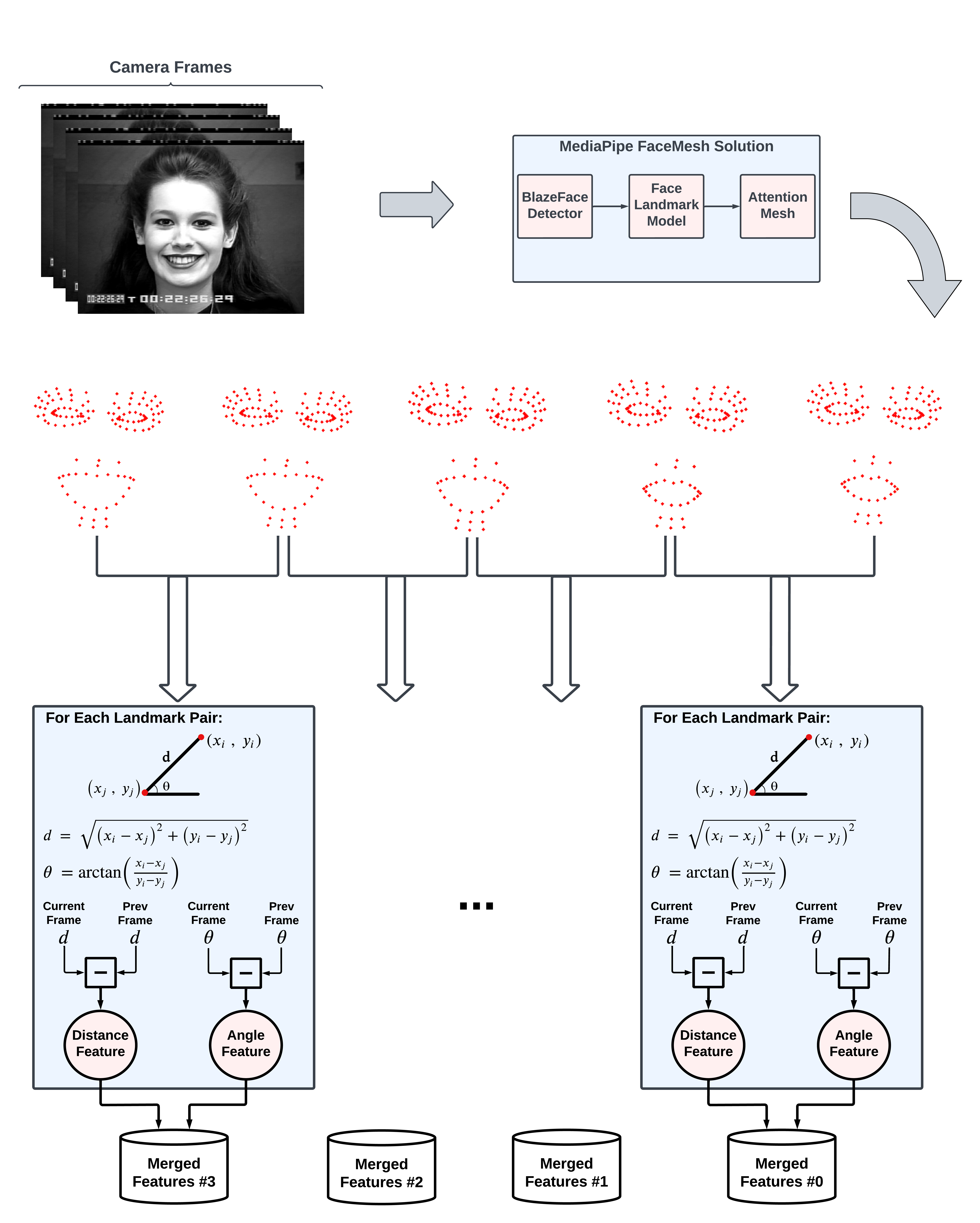}
  \captionsetup{format=hang}
  \caption{Feature creation algorithm flow. Sequential camera frames are processed by the MediaPipe FaceMesh algorithm to extract landmark coordinates. Selected frames' facial landmarks are used to create features by calculating Euclidean distance and angle between current and previous frames' landmarks. These two features are concatenated to create feature vector. Images of the subject are taken from CK+ dataset.}
  \label{fig:algo_flow_macro}
\end{figure}

The first step of the algorithm shown in Figure \ref{fig:algo_flow_macro} is to calculate facial landmarks of each camera frame. Camera frames are sequentially fed into FaceMesh algorithm, and resulting landmark coordinates are stored to be processed by the feature creation algorithm in the second step. The entire sequence of camera frames capturing facial expressions is reduced to five key frames representing the transition from a neutral expression to the apex of the expression. Facial landmarks of the selected five frames are input to the feature creation step. The feature creation algorithm generates all features belonging to the current frame by calculating the Euclidean distance and angle of each landmark pair for the current frame and previous frame. Equations 1 and 2 show the calculation of Euclidean distance and angle, respectively, for two landmark points, i and j. For each landmark pair, the calculated distance and angle values of the current frame are subtracted from the respective values of the previous frame. Resulted values give the distance and angle features of that landmark pair. 


In the real-time experiment, the algorithm's first step is modified to process each individual frame, using a sliding window buffer that holds only five frames. The same feature extraction algorithm is then applied to these frames.

\par

\begin{equation}\label{eq:distance}
d = \sqrt{(x_i - x_j)^2 + (y_i - y_j)^2}
\end{equation}

\begin{equation}\label{eq:theta}
\theta = \arctan{\left(\frac{x_i-x_j}{y_i-y_j}\right)}
\end{equation}

By default, landmark pair count is calculated by finding number of two combinations \(C(n,2)\) for total number of facial landmarks \(n\). This count can be reduced by grouping facial landmarks and calculating two combinations inside each group and combining them. Grouping landmarks ensures that there will be no landmark pair that has two landmark points belongs to two different groups.
$$C(n,2) > Unique(C(a,2) + C(b,2) + C(c,2) + C(d,2) + C(e,2))$$
where,
\begin{align*}
n &= \text{total landmark count} \\
a,b,c,d,e  &=   \text{landmark counts for 5 different groups} \\   
a,b,c,d,e  &<   n
\end{align*}
\par

In this study, landmark grouping based on facial action coding system is implemented. This grouping is similar to the one presented in the paper by Buhari et. al. \cite{buhari2020facs}. Table \ref{tab:emo_cat_map} and \ref{tab:cat_au_map} show the implemented categories and respective action units (AU) that are represented by muscles residing at shown landmark positions. Action units 1,2,3,4 and 5 are activated by facial muscles in eye and eyebrow regions so category 1 is created with landmark points on those regions. For action unit 6 eye and nose regions are selected as category 2. For action units 7 and 9, category 3 is created that consists of eye, eyebrow, and nose landmarks. For action units 12,14,15,16,23 and 26, category 4 is created that consists of nose, mouth, and lower jaw landmarks. Lastly action unit 20 consists of landmarks present in eye nose and mouth regions and category 5 is created. 

\begin{table}[htbp]\centering
    \caption{Relation between emotion, action units and category mapping.}
    \begin{tabular}{lclcl}
        \toprule
         \textbf{Emotion} & & \textbf{Action Units} & &\textbf{Required Categories}
        \\
        \midrule
        Anger & &  4, 5, 7, 23 & & cat 1, 3, 4 \\
        Contempt & &  12, 14 & & cat 4 \\
        Disgust & &  9, 15, 16 & & cat 3, 4 \\
        Fear & &  1, 2, 4, 5, 7, 20, 26 & & cat 1, 3, 4, 5 \\
        Happiness & &  6, 12 & & cat 2, 4 \\
        Sadness & &  1, 4, 15 & & cat 1, 4 \\
        Surprise & &  1, 2, 5, 26 & & cat 1, 4 \\
        \bottomrule
    \end{tabular}
    
    \label{tab:emo_cat_map}
\end{table}

\begin{table}[h]\centering
    \caption{Landmark grouping categories. Dividing facial landmarks into meaningful categories reduces total landmark pair counts that will be processed by feature creation algorithm.}
    \begin{tabular}{lcll}
        \toprule
         \textbf{Category} & & \textbf{Region} & \textbf{Action Units}
        \\
        \midrule
        cat 1
        & &
        \begin{minipage}{.2\textwidth}
            Left Eye \\
            Left Eyebrow \\
            Right Eye \\
            Right Eyebrow
        \end{minipage} &  1, 2, 3, 4, 5\\
        \midrule
        cat 2
        & &
        \begin{minipage}{.2\textwidth}
            Left Eye \\
            Right Eye \\
            Nose
        \end{minipage} &  6\\
        \midrule
        cat 3
        & &
        \begin{minipage}{.2\textwidth}
            Left Eye \\
            Left Eyebrow \\
            Right Eye \\
            Right Eyebrow \\
            Nose
        \end{minipage} &  7, 9\\
        \midrule
        cat 4
        & &
        \begin{minipage}{.2\textwidth}
            Nose \\
            Mouth \\
            Lower Jaw
        \end{minipage} &  12, 14, 15, 16, 23, 26\\
        \midrule
        cat 5
        & &
        \begin{minipage}{.2\textwidth}
            Left Eye \\
            Right Eye \\
            Nose \\
            Mouth
        \end{minipage} &  20\\
        \bottomrule
    \end{tabular}
    
    \label{tab:cat_au_map}
\end{table}

\subsection{Classification Algorithm}

\begin{figure}[htbp]
  \centering
  \includegraphics[width=\textwidth]{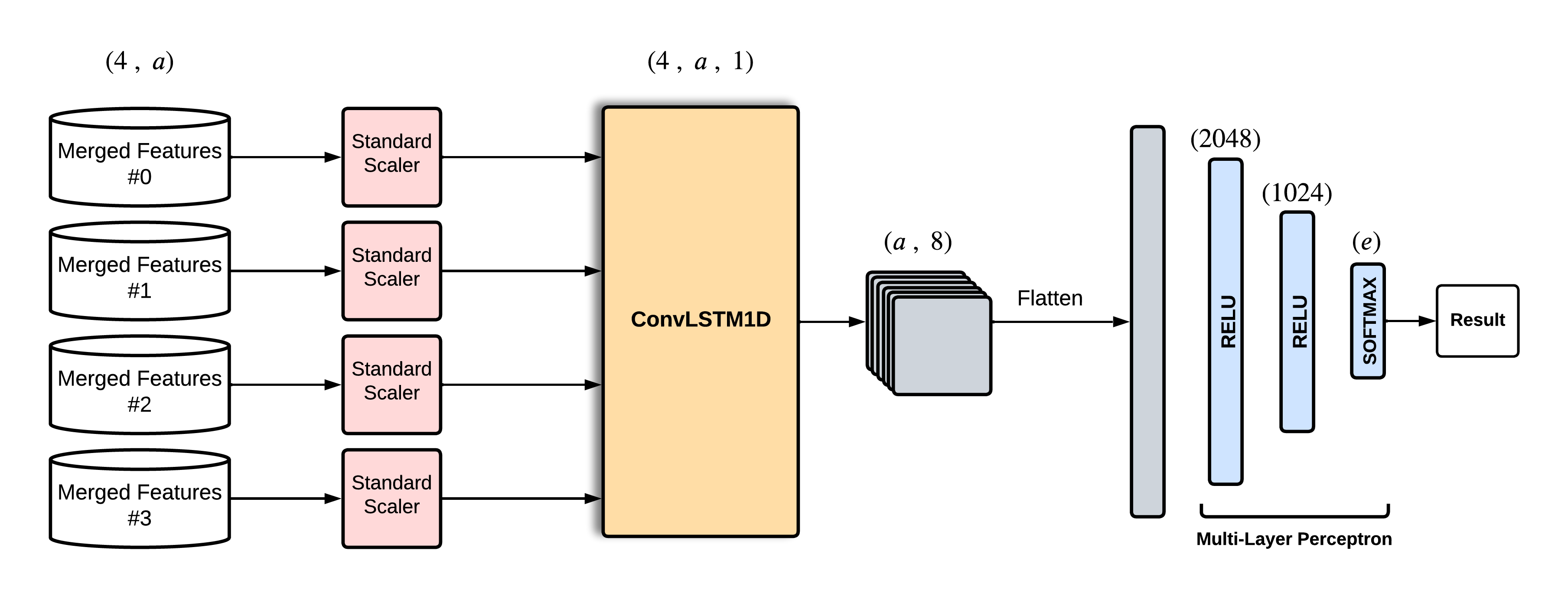}
  \captionsetup{format=hang}
  \caption{Classification algorithm for macro-expression experiments. First \(N-1, A\) shaped array that holds feature vectors for whole sequence is scaled using standard scaler. Scaled 1D data is converted to image format that will be feed into ConvLSTM1D block. Output of ConvLSTM1D block is flattened and data is classified using multi-layer perceptron layers. Where, \(e\): emotion count, \(a\): feature count.}
  \label{fig:algo_class_micro}
\end{figure}

\par
In this study, Multilayer Perceptron is selected as a classifier. In a study by Alvarez et al., it is concluded that for facial emotion recognition tasks, Multilayer Perceptron is the classifier that achieves the highest accuracy compared to SVM, Naïve Bayes, Decision Tree, Random Forest, and AdaBoost \cite{alvarez2018facial}. Figure \ref{fig:algo_class_micro} shows the classification part of the proposed method after features are created.
\par
To extract information from temporal domain ConvLSTM1D is used. After features are created and stored in \(N-1, A\) shaped array where \(N\) is the frame count and A is the total feature count, data is scaled before classification. Standard scaler is used for this purpose.
Scaled 1D features are fed into ConvLSTM1D block with a kernel size 1 and filter size 8. Output of the ConvLSTM1D is flattened and dense layers of 2048 and 1024 neurons respectively are used in multi-layer perceptron. Final classification layer has neuron count which is equal to emotion count to be classified and softmax activation function is used. 
\par
ConvLSTM is a type of neural network architecture that combines convolutional layers with LSTM (Long Short-Term Memory) layers. It is commonly used for sequence prediction tasks, such as video and image sequence processing, where both spatial and temporal dependencies need to be modeled. In ConvLSTM, the input data is processed by convolutional layers to capture spatial features, and then the output of the convolutional layers is passed to LSTM layers, which capture temporal dependencies. The LSTM layers maintain an internal state that enables them to capture long-term dependencies in the sequence \cite{shi2015convolutional}.

\par
LSTM networks are a special kind of Recurrent Neural Networks (RNN) that have feedback connections, allowing them to process sequences of data \cite{hochreiter1997long}. They are capable of learning long-term dependencies, which makes them particularly effective for many sequential data tasks. The ConvLSTM is a type of LSTM that has convolutional structure in both the input-to-state and state-to-state transitions \cite{shi2015convolutional}. This makes it uniquely suited to handle two-dimensional spatial data over time. Each unit of a ConvLSTM network maintains a cell state and multiple gating units, including an input gate, a forget gate, and an output gate, which control the flow of information into and out of the cell. The convolution operation is applied in the state transition and the gate activations, which allows the ConvLSTM to effectively capture the spatial dependencies in the data. The operations utilized within LSTM are reconfigured for ConvLSTM, as indicated in equations \ref{eq1}-\ref{eq5}. 

\begin{figure}[htbp]
  \centering
  \includegraphics[width=0.85\textwidth]{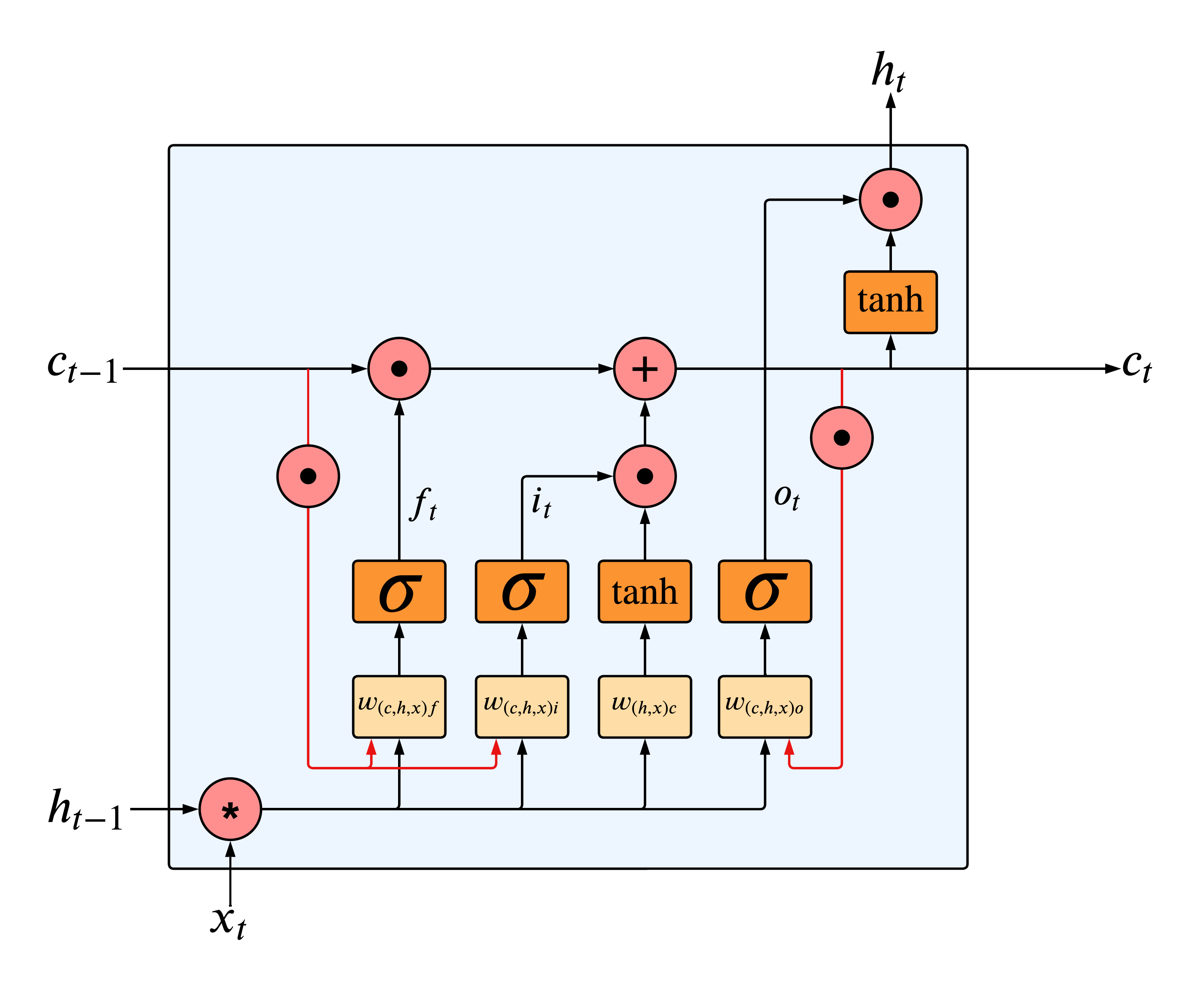}
  \caption{Diagram illustrating the internal structure of a ConvLSTM cell. The figure depicts the convolutional operations applied to the input and hidden states, showcasing the processes of input modulation, forget gate, input gate, and output gate. These components work together to capture spatiotemporal features in sequential data.}
  \label{fig:convlstm_inner}
\end{figure}

In this context, the symbols \("*"\) and \("o"\) correspond
to the convolution operation and the Hadamard product, respectively. "x" stands for the input vector, which is the data that the network is receiving at a given time step. "h" stands for the hidden state vector, which represents the internal state of the network at a given time step. It serves as the memory of the network, allowing it to keep track of relevant information from previous time steps and make predictions based on that information. "c" stands for cell state which allows the network to selectively remember or forget information based on its relevance to the current task. Figure \ref{fig:convlstm_inner} shows the inner structure of ConvLSTM cell. The use of ConvLSTM in facial expression analysis allowed us to effectively capture the spatial and temporal dependencies in our data and provide valuable insights into the problem.

\begin{equation}
i_t = \sigma(W_{xi} * x_t + W_{hi} * h_{t-1} + W_{ci} \circ c_{t-1} + b_i)
\label{eq1}
\end{equation}

\begin{equation}
f_t = \sigma(W_{xf} * x_t + W_{hf} * h_{t-1} + W_{cf} \circ c_{t-1} + b_f)
\label{eq2}
\end{equation}

\begin{equation}
c_t = f_t \circ c_{t-1} + i_t \circ \tanh(W_{xc} * x_t + W_{hc} * h_{t-1} + b_c)
\label{eq3}
\end{equation}

\begin{equation}
o_t = \sigma(W_{xo} * x_t + W_{ho} * h_{t-1} + W_{co} \circ c_t + b_o)
\label{eq4}
\end{equation}

\begin{equation}
h_t = o_t \circ \tanh(c_t)
\label{eq5}
\end{equation}

\par
Standard scaling is a preprocessing step in machine learning that scales the data so that each feature has zero mean and unit variance. This is important because many algorithms assume that the data is normally distributed with zero mean and unit variance. Standard scaling can be applied to both training and test data and is particularly useful when dealing with features with different scales or units \cite{raju2020study}. It is implemented in many popular machine learning libraries, such as Scikit-learn in Python \cite{pedregosa11a}.
\begin{equation}\label{eq:standardscale}
z = \frac{x - u}{s}
\end{equation}
where:
\begin{align*}
 u & = \text{mean of the training samples} \\
 s & =  \text{standard deviation of the training samples} \\
\end{align*}

\section{Results and Discussions}\label{sec4}
In order to implement facial expression analysis methods in real-world scenarios, they need to satisfy several criteria. Among these, processing time emerges as a crucial consideration for real-time applications. The cumulative time required for preprocessing, feature creation, and classification stages should not surpass the interval between two consecutive frames captured by the camera. Moreover, models that incorporate the temporal dynamics of facial features can yield more robust models compared to those that rely solely on spatial features. Relying solely on a single static moment may lead to misinterpretations, particularly when a person’s neutral state closely resembles a specific emotional expression. Facial expressions are dynamic in nature and continuously change over time. Thus, analyzing the entire sequence of expressions is more appropriate. This enables us to examine the onset, apex, and offset phases of emotion within the temporal domain, facilitating accurate detection and labeling of the entire sequence with the corresponding emotional tag. This approach is more precise and can provide valuable insights into the dynamics of facial expressions. In our study, we adopt a sequential approach to facial emotion recognition tasks, an advancement over traditional static methods. Our technique is designed to compare and differentiate all features of frames within an expression sequence from their previous states.

\subsection{Experiments with Datasets Individually}
\begin{figure}[htbp]
  \centering
  \includegraphics[width=\textwidth]{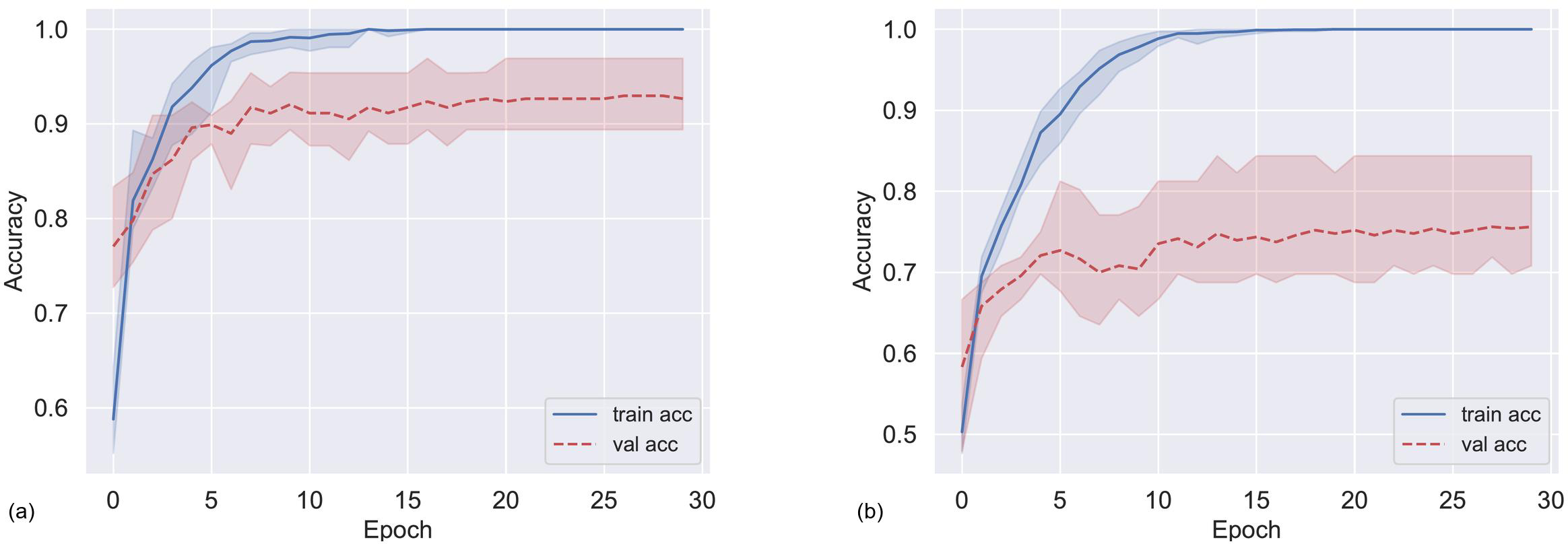}
  \captionsetup{format=hang}
  \caption{Accuracy vs epoch graphs for (a) CK+ and (b) Oulu-CASIA datasets trained using 61 landmarks with AU grouping. Shaded area shows min-max regions of each fold of 5-fold cross validation while bold lines shows average of them.}
  \label{fig:acc_vs_epoch}
\end{figure}

In order to validate our model, 5-Fold cross validation was used to calculate average accuracies for CK+, Oulu-CASIA NIR \& VIS and MMI datasets. Five-fold cross-validation is a method for evaluating machine learning models. The dataset was split into five subsets, or folds. The model is then trained and evaluated five times, with each iteration using a different fold as the validation set and the remaining four folds as the training set. Performance metrics were calculated for each iteration, and the average performance was determined by averaging these metrics across all folds. This approach provides a more reliable estimate of the model's performance and helps assess its generalization ability. Figure \ref{fig:acc_vs_epoch} shows the accuracy vs epoch graph for two selected datasets. These graphs are useful for monitoring the model's performance during training, analyzing training behavior, fine-tuning hyperparameters, and comparing results. 
\par
\begin{figure}[htbp]
  \centering
  \includegraphics[width=\textwidth]{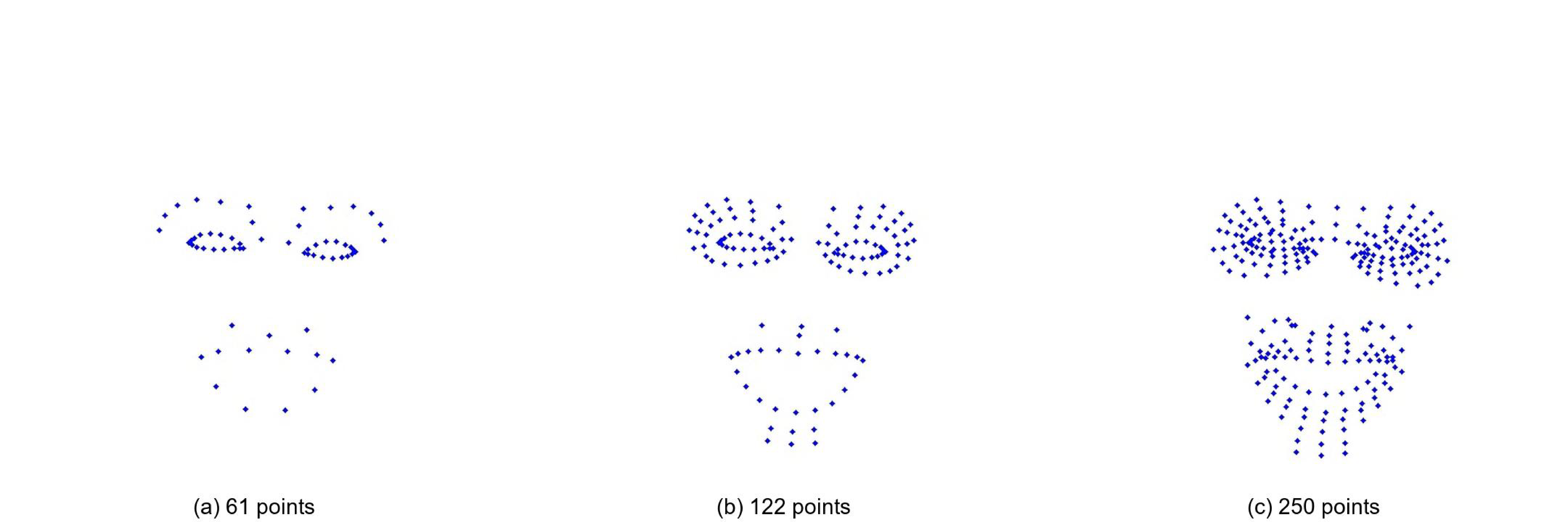}
  \captionsetup{format=hang}
  \caption{Selected facial landmark sets used in macro-expression experiments. 61, 122 and 250 landmark points are selected manually from 478 facial landmark. Images of the subject are taken from CK+ dataset.}
  \label{fig:used_landmarks}
\end{figure}

Different experiments are conducted to expose correlation between accuracy and the number of facial landmarks, along with the number of created features. To find out the relation between accuracy and the number of facial landmarks, three different preset landmark counts are defined as 61, 122, and 250. Those landmarks seen in Figure \ref{fig:used_landmarks} are selected manually from 478 landmarks of FaceMesh output. Landmarks are selected based on facial muscle locations on the face according to action units (AU). Main action units that are used to recognize emotions located on eye, eyebrow, mouth, nose, and chin regions on the face. To find out the relation between accuracy and the number of created features, a feature selection algorithm based on FACS by Buhari et al. \cite{buhari2020facs} is utilized to reduce the number of generated features. So, in total six experiments are conducted for each dataset. 
\par
\begin{figure}[htbp]
  \centering
  \includegraphics[width=\textwidth]{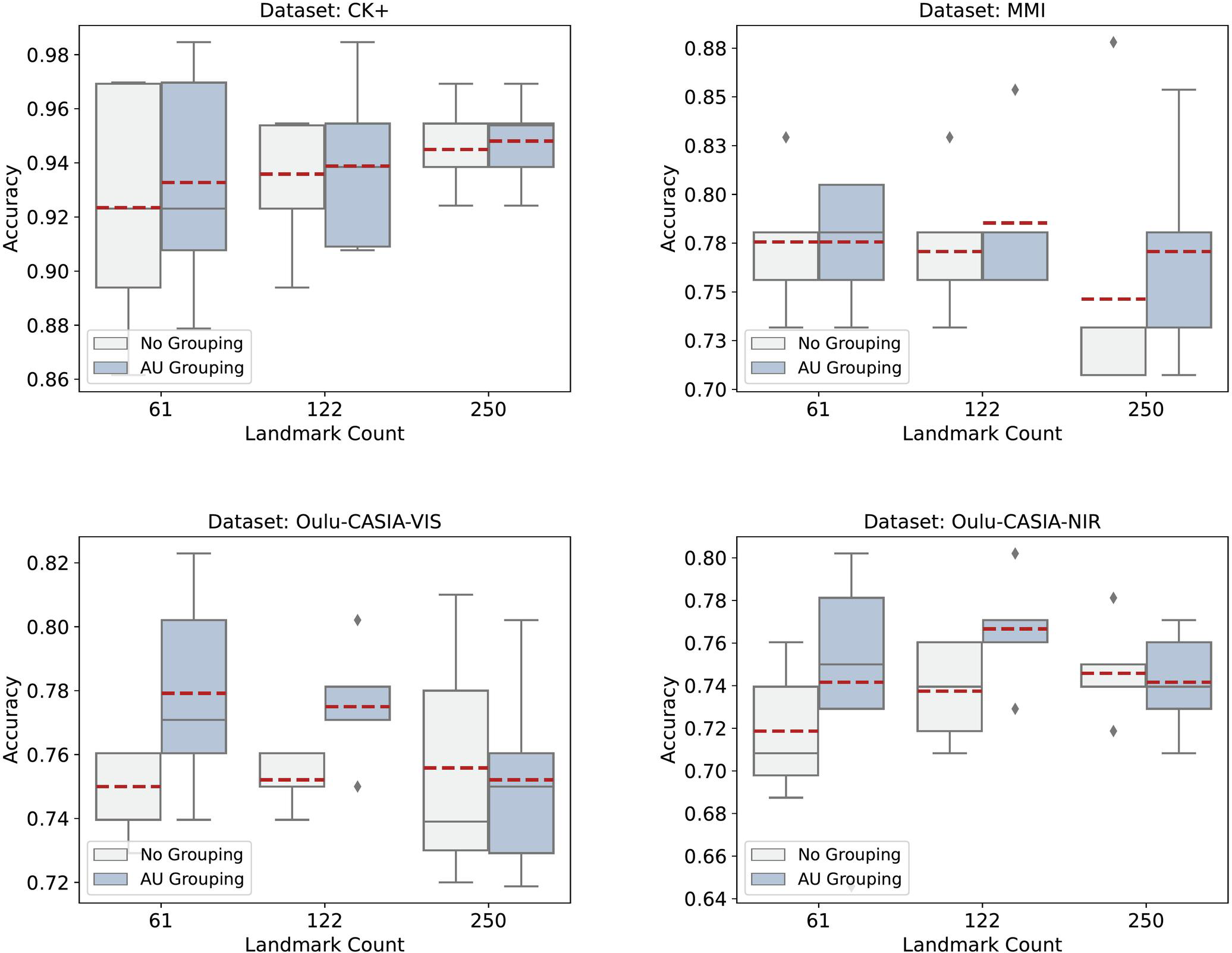}
  \captionsetup{format=hang}
  \caption{Accuracy box-plots of datasets used in macro-expression experiments. Results for all six experiments that are combinations of 61, 122, 250 point landmarks with AU grouping and without any grouping. Red dashed line shows mean value of accuracy for 5-fold cross validation.}
  \label{fig:acc-box-plot}
\end{figure}

In Figure \ref{fig:acc-box-plot}, box plots of accuracies for every dataset are shown. For each plot, results for all six experiments, respectively 61, 122, and 250 points landmarks with AU grouping and without any grouping, are visible. The box plots display the mean accuracy values of all five folds, indicated by a red dashed line. The minimum and maximum values excluding outliers are represented by grey lines below and above the boxes, as well as grey dots if they are outliers. It can be deduced from conducted experiments that grouping facial landmarks based on FACS usually offers better results since it acts as a feature selection method that selects prominent features among all. Also, it can be stated that increasing landmark counts does not significantly increase the accuracies, and sometimes it even has negative effects like it is observed in the MMI dataset. The CK+ dataset achieved its highest mean accuracy of 93\% in experiment 250 landmarks with AU grouping. For the MMI dataset, the highest mean accuracy of 68\% was recorded in experiment 61 landmarks without grouping. In the case of the Oulu-CASIA VIS dataset, the results showed that experiment 250 landmarks without grouping had the highest mean accuracy of 79\%. However, the Oulu-CASIA NIR experiment yielded slightly lower results compared to the VIS version. The experiment with 250 landmarks and AU grouping had the highest mean accuracy of 77\%.

\par
\begin{figure}[htbp]
\centering
  \includegraphics[width=\textwidth]{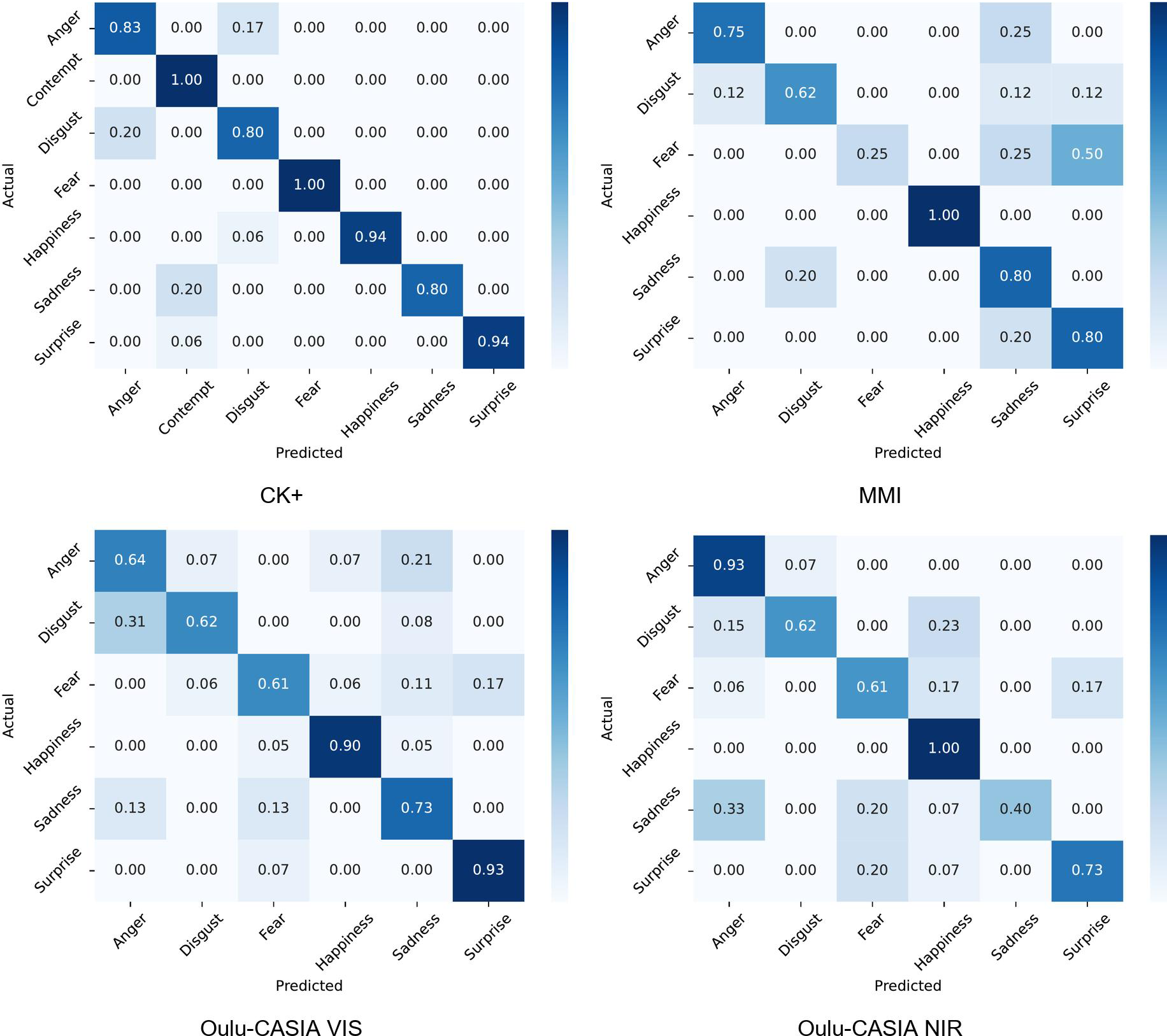}
  \captionsetup{format=hang}
  \caption{Confusion matrices for facial expression recognition on (a) CK+, (b) MMI, (c) Oulu-CASIA VIS, and (d) Oulu-CASIA NIR datasets. Results show average performance of 5-fold cross-validation. Darker colors and higher numerical values indicate better accuracy.}
  \label{fig:conf_matrix_macro}
\end{figure}

It’s helpful to use confusion matrices to better understand how accurately emotions are predicted and which emotions are often confused with others. In Figure \ref{fig:conf_matrix_macro}, it can be observed that for CK+ dataset, contempt is commonly confused with fear and sadness, while fear is often confused with surprise and sadness for the MMI and Oulu-CASIA NIR datasets. In contrast, sadness is the most difficult emotion to predict accurately for the Oulu-CASIA VIS dataset. Furthermore, happiness and surprise emotions tend to have the highest prediction accuracy across all datasets.

\par
We used MediaPipe’s FaceMesh solution to extract facial landmarks, which has superior performance on GPUs, making it suitable for real-time tasks. The time required to extract all landmarks with our hardware setup is measured as 5.6 ms, regardless of the landmark count used in the experiment. This is significantly faster than dlib’s landmark detection algorithm, which takes 100 ms to process. In the second phase, the time taken to create features depends on selected landmark point pairs. Table \ref{tab:process_times_landmarks} provides the measured time to create features for different landmark point counts and categories. It’s worth noting that using a GPU to create features is 12 times faster than using a CPU with our hardware setup. The fastest processing time recorded is 0.45ms, which was achieved by creating features for 61 landmarks with AU grouping. For 61 landmark points with AU grouping, the total processing time, from capturing camera frames to create all features for each frame, is 6.05 ms on Nvidia RTX 3060 with 8.6 compute capability. This value indicates that the method can support approximately 165 fps video processing in real-time. This means that the system can analyze video frames at a high speed, allowing for efficient and timely recognition of facial expressions. Additionally, the method was evaluated using the Oulu-CASIA dataset, which consists of frames captured under both visible light (VIS) and near-infrared light (NIR) conditions. This evaluation aimed to demonstrate the robustness of the method across different ambient light conditions without the need for additional preprocessing. By successfully classifying facial expressions under varying lighting conditions, the method proves its versatility and suitability for real-world applications. Prediction time of pretrained model for 61 landmark AU grouping input type is measured as 40ms.

\begin{table}[htbp]
    \centering
    \caption{Comparison of processing times to create features for GPU and CPU execution. Full means using all landmark pairs and AU means using selected landmark pairs based on FACS based method.}
    \begin{tabular}{l *{9}{c}}
        \toprule
         & &
        \multicolumn{2}{c}{61 LM} & &
        \multicolumn{2}{c}{122 LM} & &
        \multicolumn{2}{c}{250 LM} \\
        \cmidrule(){3-4} \cmidrule(lr){6-7} \cmidrule(l){9-10}
         & & Full & AU & & Full & AU & & Full & AU \\
        \midrule
        GPU (ms) & & 0.65 & 0.45 & & 2.60 & 1.65 & & 11.01 & 8.08 \\
        CPU (ms) & & 7.74 & 5.17 & & 31.13 & 21.53 & & 130.67 & 91.20 \\
        \bottomrule
    \end{tabular}
    
    \label{tab:process_times_landmarks}
\end{table}

\begin{table}[htbp]\centering
    \caption{Comparison of accuracy values of different methods in the literature which use only geometric features}
    \begin{tabular}{ccccc}
        \toprule
        \textbf{Paper} &&
        \multicolumn{3}{c}{\textbf{Accuracy(\%)}} \\
        \cmidrule(lr){3-5}
        && \textbf{CK+} & \textbf{Oulu-CASIA} & \textbf{MMI} \\
        \midrule
        Aikyn et al.\cite{Aikyn2024} && 95.12 & - & - \\
        Jung et al. (DTGN)\cite{jung2015joint} && 92.35 & 74.17 & 59.02 \\
        Choi et al. (2D LFM)\cite{choi2020facial} && 92.60 & - & - \\
        Qiu et al. (multiple-origin)\cite{qiu2019facial} && 92.00 & - & - \\
        Raj et al.\cite{rohith2020facial} && 89.00 & - & - \\
        Álvarez et al.\cite{alvarez2018facial} && 89.00 & - & - \\
        \textbf{Proposed method} && 93.00 & 79.00 & 68.00 \\
        \bottomrule
    \end{tabular}
    
    \label{tab:acc_compare}
\end{table}

\par
The accuracies of other geometric-based methods in the literature, which operate on the datasets used in this study, are presented in Table \ref{tab:acc_compare}. While there are not many geometric-based methods that have been tested on the Oulu-CASIA and MMI datasets, our proposed method surpasses the performance of the mentioned methods in terms of recognition accuracy. 
\subsection{Composite Dataset Experiments}
In this section, a single composite dataset is created by merging multiple datasets, and experiments are conducted using this composite dataset. The aim of these experiments is to show that the model learns features that are generalizable across various datasets, rather than being specific to the trained dataset. The first experiment involves training the proposed method on three datasets and validating it on a fourth dataset that is not included in the composite dataset. The second experiment combines all the datasets and performs internal validation using a 5-fold cross-validation approach. 
\par
In the first experiment, the CK+, Oulu-CASIA NI, and Oulu-CASIA VIS datasets are combined by focusing on six basic emotions: anger, disgust, fear, happiness, sadness and surprise, which are common to all datasets. Emotions such as contempt and others were excluded to facilitate merging the datasets into a single composite dataset. The composite dataset contains the following sequence counts for each emotion: anger (205), disgust (219), fear (185), happiness (229), sadness (188), and surprise (243). The resulting composite dataset was then used to train the proposed model, which was subsequently validated using the MMI dataset, not included in the composite dataset. The validation accuracy on the MMI dataset was 69.70\%, slightly higher than the 68\% achieved during the individual training on the MMI dataset. 
\par
In the second experiment, all datasets, including CK+, Oulu-CASIA NI, Oulu- CASIA VIS, and MMI, were combined to create a composite dataset. This composite dataset includes 476 sequences for anger, 500 sequences for disgust, 426 sequences for fear, 540 sequences for happiness, 440 sequences for sadness, and 566 sequences for surprise emotions. The model is validated using a 5-fold cross-validation approach, resulting in an accuracy of 82.10\%.
\par
The results show that the model can learn features from subject images captured in various environmental conditions. Mixing different subjects in these conditions does not affect the model's performance, indicating its robustness to environmental variations.
\subsection{Real-time Implementation}
Real-time facial expression recognition has gained significant importance in various fields due to its potential applications in understanding human emotions, enhancing human-computer interaction and enabling affective computing systems. The ability to detect and interpret facial expressions in real-time opens up possibilities for emotion-aware technologies, intelligent surveillance systems and personalized user experiences accurately and swiftly. 
\begin{figure}[htbp]
  \centering
  \includegraphics[width=0.8\textwidth]{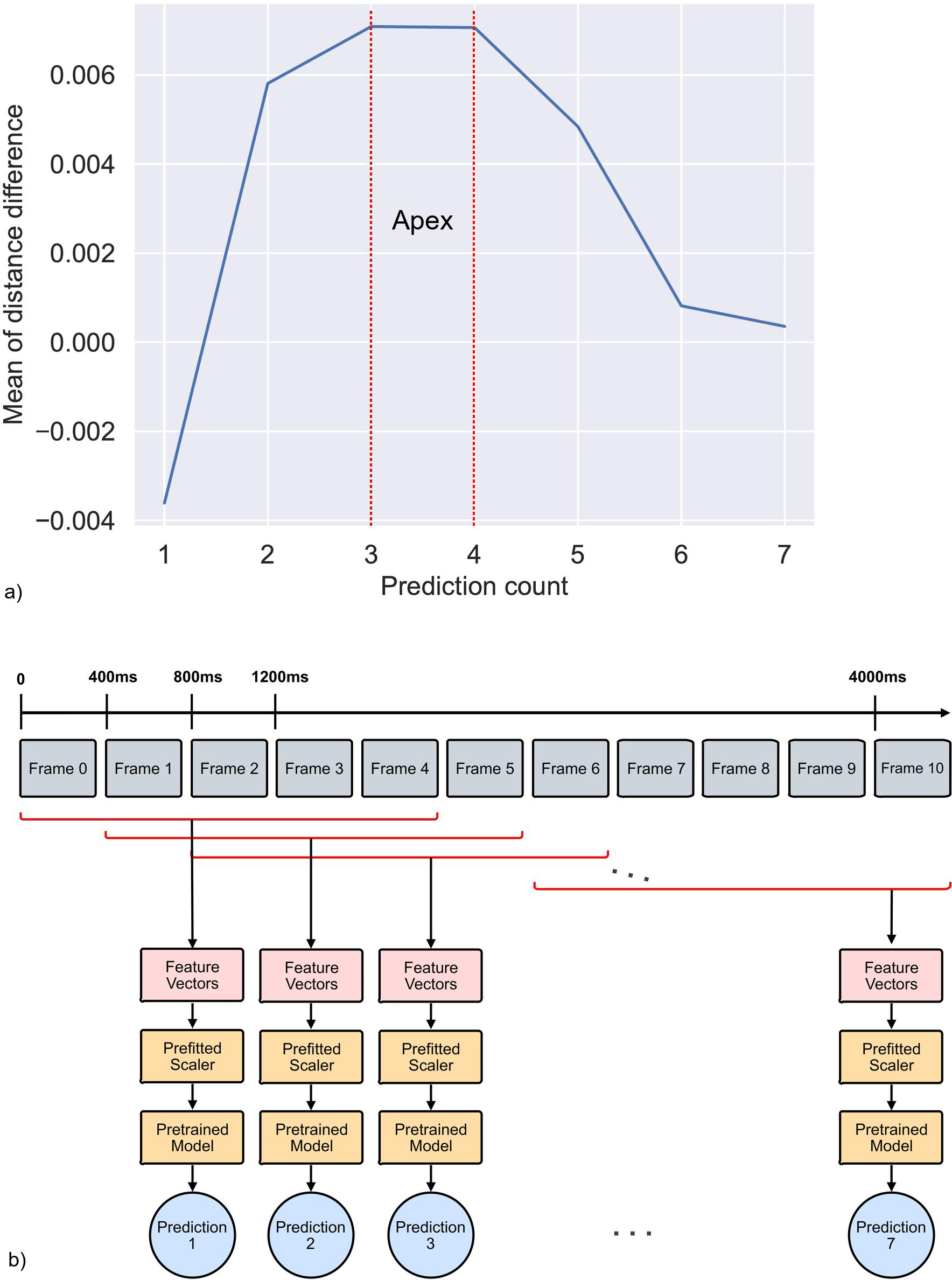}
  \captionsetup{format=hang}
  \caption{a) Real-time detection of facial expression phases. Y-axis shows mean Euclidean distance feature difference between first and last frames in the buffer; x-axis shows prediction count. Apex region corresponds to peak difference and highest prediction accuracy  b) Real-time implementation flow chart for facial expression recognition. Frames are captured every 400ms and processed in overlapping sequences of 5 frames. Feature vectors are extracted from each sequence, scaled using a prefitted scaler, and fed into a pretrained model for emotion prediction. This sliding window approach enables continuous real-time analysis of facial expressions
  }
  \label{fig:dist_over_time}
\end{figure}
By accurately recognizing facial expressions in real-time, we can enable adaptive interfaces, emotion-driven assistive technologies and immersive virtual experiences that respond to users' emotional states.

\par
The model was trained using the composite dataset created in the first experiment in section 4.2. After training, the model and scaler were saved. The pretrained model was used to predict the emotion class of the validation data, while the prefitted scaler was used to scale the features of the validation data. The model’s input consisted of 4 feature vectors was created from 5 frames. Therefore, frames were captured every 400ms and appended to a buffer. When the buffer reached 5 frames, feature vectors were created and input into the model for prediction. After each prediction, the first element of the buffer was deleted, and all other elements were shifted to make a space for a new frame. 
\par
Apex is the most intense moment that an emotion can be observed in the face. In order to detect facial macro expression in real-time, determining the apex region is critical since the accuracy of prediction is higher than in other regions. In Figure \ref{fig:dist_over_time} it can be observed that by tracking the mean value of distance features of frames, onset, apex, and offset regions can be detected. Since coordinates of landmarks are expressed as pixels, the value shows pixel difference of first and last frames’ features in the buffer.
\par
\begin{figure}[htbp]
  \centering
  \includegraphics[width=\textwidth]{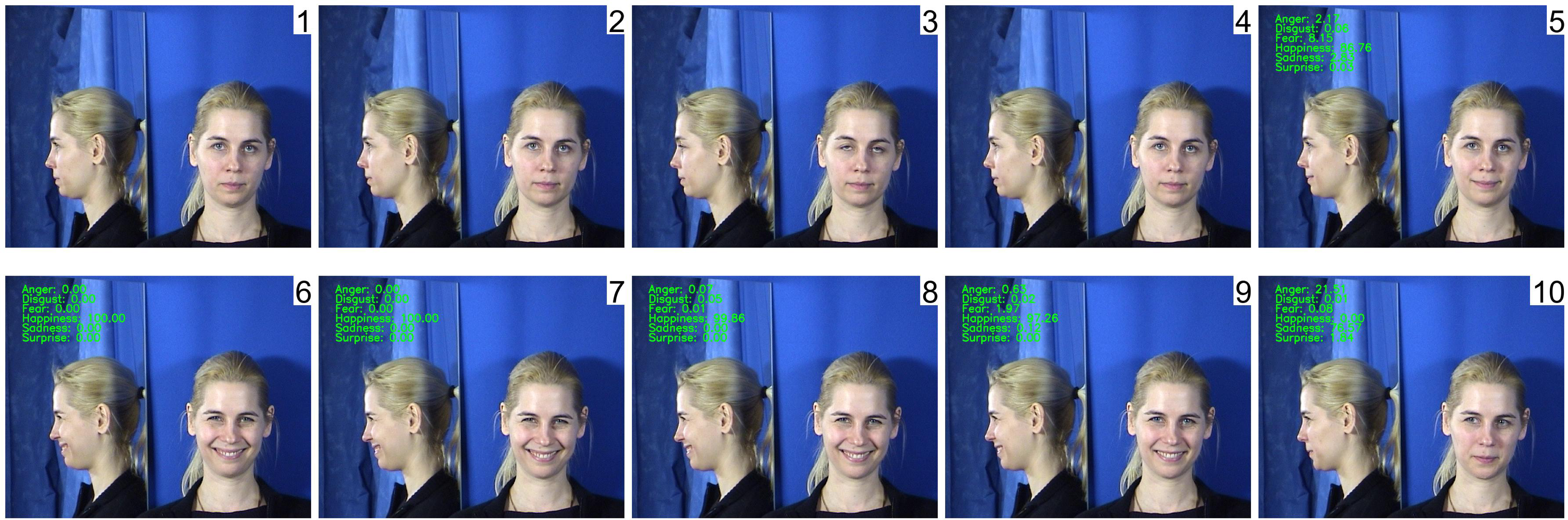}
  \captionsetup{format=hang}
  \caption{Real-time facial expression recognition demonstrating the model's ability to track emotional progression. First frame is taken in neutral phase. First 5 frames shows onset phase where facial expression is started. At frame number 5, buffer is filled and first prediction is done. Prediction accuracy increases throughout frames 5 to 9, reaching 100.00\% at apex. Image sequence is taken from MMI dataset.}
  \label{fig:real-time-predict}
\end{figure}

To validate the real-time performance of the model, a specific video displaying a happiness macro expression, starting from a neutral state, was selected. This video sourced from the MMI dataset, was not included in the model’s training phase. The results of this real-time validation process are shown in Figure \ref{fig:real-time-predict}. 
\par
The real-time prediction results illustrate the progression of emotion throughout the video frames. Initially, the model begins to recognize the emerging emotion using the first 5-frame sequence. After frame 5, where the emotion is fully expressed, the model predicts the emotion with very high accuracy until the last frame. This indicates that the model successfully captures the features and patterns specific to the expressed emotion, leading to highly accurate predictions.

\par
In conclusion, this paper presented experiments on facial macro expression recognition using various datasets. The performance of the proposed methodology was evaluated using a 5-fold cross-validation approach. The results indicate promising accuracy rates for the composite dataset, achieving an accuracy of 82.10\%. Specifically, the CK+ dataset exhibited a high accuracy of 93\%, while the MMI dataset achieved an accuracy of 68\%. Additionally, the Oulu-CASIA VIS dataset and Oulu-CASIA NIR dataset achieved accuracies of 79\% and 77\%, respectively. These findings demonstrate the effectiveness of our approach in accurately recognizing facial macro expressions across different datasets. Furthermore, we successfully demonstrated real-time implementation based on the model created using the composite dataset. These outcomes highlight the potential practical applications of our methodology in real-world scenarios requiring real-time facial macro expression recognition.

\section{Conclusion}\label{sec5}
In recent years, facial expression recognition has become a critical area of research in artificial intelligence. This study proposes a deep learning-based sequential macro-expression recognition method. Facial landmarks are detected using MediaPipe’s FaceMesh solution, which is significantly faster than the popular dlib algorithm. Geometric features are created from facial landmarks by considering the differences in Euclidean distance and angle features from their previous states in the sequence. It is shown that by tracking the mean value of these differences over time, the onset, apex, and offset phases of an emotion can be detected.
In our experiments, it was observed that increasing the number of landmarks does not necessarily improve accuracy and can sometimes have negative effects. Experiments with the FACS based landmark grouping method show that selecting useful features using a feature reduction algorithm often increases classification accuracy. With the proposed method, competitive mean accuracy values were achieved among landmark-based methods in the literature using a 5-fold cross-validation technique. The proposed method was tested with CK+, Oulu-CASIA VIS \& NIR, and MMI datasets, achieving accuracy results of 94\%, 78\%, 77\%, and 79\%, respectively. Composite dataset experiments involved merging multiple datasets to observe the generalization of the proposed model. The real-time implementation of the model was validated using a video displaying the emotion of happiness, achieving remarkable prediction accuracy. These results demonstrate the model’s ability to generalize across datasets and accurately predict emotions in real-time scenarios.
Despite the advancements, achieving a robust, accurate, and real-time solution for facial expression recognition remains an ongoing challenge. Given the escalating prevalence of human-computer interaction, it is foreseeable that many applications across diverse areas will require capabilities to detect human emotions. To be viable for real-world adoption, the proposed system must not only be fast but also resilient against various challenges, including changes in illumination, face rotation, facial accessories, and other potential distortion factors. Our study paves the way for continued exploration in this field, contributing to the pursuit of a versatile, precise, and user-adaptive solution for facial expression recognition.

\section*{Statements and Declarations} \label{declarations}

\textbf{Competing interests} The authors have no relevant financial or non-financial interests to disclose.

\noindent{\textbf{Ethics approval} All data used in this manuscript are publicly available datasets and do not involve personal privacy data of living individuals, and no ethical factors are involved.}

\noindent{\textbf{Data availability} The datasets supporting the findings of this study are publicly available.}\\
CK+ dataset can be requested at http://www.jeffcohn.net/Resources. \\
Oulu-CASIA dataset can be requested at https://www.oulu.fi/en/university/faculties-and-units/faculty-information-technology-and-electrical-engineering/center-for-machine-vision-and-signal-analysis. \\
MMI dataset can be requested at https://mmifacedb.eu/.

\clearpage

\bibliographystyle{iopart-num}
\section*{References}
\bibliography{sn-bibliography}

\end{document}